\documentclass{article}

\usepackage[final,dblblindworkshop]{neurips_2026}

\usepackage[utf8]{inputenc}
\usepackage[T1]{fontenc}

\usepackage{amsmath,amssymb,amsfonts}
\usepackage{booktabs}
\usepackage{graphicx}
\usepackage{microtype}
\usepackage{hyperref}
\usepackage{url}

\graphicspath{{figures/}}

\workshoptitle{AI for Drug Discovery: Bridging the Translation Gap}
\title{Geometric Relevance Is Not Sufficient: Evaluating Focused Post-Training Supervision in Diffusion Docking}
\author{%
  Michael Brown \quad Maryam Rahimimovassagh \quad Mehdi Yazdani-Jahromi \\
  \textbf{Sanjeda Sara Jennifer} \quad \textbf{Ivan Garibay} \\
  \normalfont University of Central Florida \\
  \normalfont Orlando, FL, USA
}

\begin{document}

\maketitle

\begin{abstract}
Diffusion-based molecular docking requires learning a corrective score field across a broad corruption distribution, yet individual pose errors can differ substantially in their structural consequences. This creates a post-training conflict between preserving broad score-field coverage and concentrating supervision on corrections that appear more relevant to docking geometry. Using DiffDock \citep{corso2023diffdock}, a fixed-budget study compares broad continuation with torsional supervision focused on Cartesian consequence or conditional target consistency, and with a broad/focused mixture. Evaluation separates fixed score fidelity, generated-pose quality, best-of-10 sampling capacity, and confidence-selected Top-1 docking performance. At $<5\,\text{\AA}$, broad continuation had the highest mean Top-1 success among trained policies, and focused or mixed supervision showed no consistent advantage under the tested budget. The starting model had better fixed score fidelity, whereas the evaluated broad-continuation model had higher mean Top-1 success with the same aggregate best-of-10 recovery. Focusing supervision on structurally consequential or more consistently specified corrections did not consistently improve returned-pose success over broad continuation.
\end{abstract}

\section{Introduction}
Accurate protein--ligand binding-pose prediction supports structure-based drug discovery by enabling interaction modeling, hit identification, and optimization \citep{zhu2024diffbindfr,alcaide2024unimoldockingv2}. DiffDock separates pose generation from confidence-based selection by sampling ligand translations, rotations, and torsions and then ranking the resulting poses \citep{corso2023diffdock,corso2024deepconfident,khiari2026agenticposesranker}. A near-native pose may therefore be generated without being selected for downstream use, so docking evaluation distinguishes structural accuracy from physical validity and interaction recovery \citep{buttenschoen2023posebusters,errington2024interactionrecovery,morehead2024dockingpotential}.

A score model learns a time-dependent corrective field over states produced by a corruption process, so its training objective averages error across examples, perturbations, and diffusion times \citep{song2021scorebased,corso2023diffdock}. Torsional errors act in Cartesian space through atoms displaced around each rotatable bond. Equal angular errors can therefore have different consequences for the ligand pose depending on the bond-axis geometry and the atoms affected \citep{corso2023diffdock,jing2022torsionaldiffusion}. Cartesian consequence and conditional target consistency provide distinct bases for focusing torsional supervision. These requirements collide during fixed-budget post-training. Broad continuation preserves coverage of the learned score-field distribution, while focused supervision concentrates gradient pressure on corrections with greater Cartesian consequence or more consistent targets \citep{song2021scorebased,corso2023diffdock}. A geometric diagnosis can identify a difficult or consequential state without specifying the corrective denoising direction required by the reverse process \citep{jing2022torsionaldiffusion,corso2023diffdock}. The decision is whether limited post-training should preserve broad coverage or redirect torsional supervision.

This study examines this decision in DiffDock through a matched five-epoch post-training experiment \citep{corso2023diffdock,corso2024deepconfident}, with supporting ablations of geometric relevance, directional action, and torsion-priority criteria detailed in the appendix. Evaluation separates fixed score fidelity from generated-pose success, best-of-10 sampling capacity, and confidence-selected Top-1 docking performance \citep{buttenschoen2023posebusters,errington2024interactionrecovery}. No focused or mixed policy showed a consistent advantage over broad continuation on the primary $<5\,\text{\AA}$ endpoints. The starting-checkpoint versus post-training EMA comparison showed higher mean Top-1 success alongside worse fixed score fidelity and unchanged aggregate best-of-10 recovery.

\section{Competing Objectives and Evaluation}
DiffDock represents a docking state as $z=(\mathbf t,R,\boldsymbol\theta)$, with translational, rotational, and torsional coordinates corrupted across diffusion time \citep{corso2023diffdock}. Prior diffusion work reallocates denoising emphasis across noise levels or timesteps \citep{song2021scorebased,karras2022edm,choi2022p2,hang2023minsnr,nichol2021improvedddpm}; here, the corrupted-state stream is fixed and torsional loss is reallocated across ligand edges using allocation masks and intrinsic or Cartesian metrics. Focused post-training does not alter the sampled complexes, perturbation stream, translation objective, or rotation objective. Instead, it changes the torsional contribution to the objective. Let $A$ denote the torsion-allocation policy and let $Q\in\{I,\widetilde G\}$ denote the torsional metric. For the corresponding torsional loss $\ell_{\mathrm{tor}}^{A,Q}$,
\begin{equation}
\begin{aligned}
L_{A,Q}(\phi)
  &= \mathbb{E}_{z\sim\mu}
  \left[\ell_{\mathrm{tr}}+\lambda_R\ell_{\mathrm{rot}}+\lambda_T\ell_{\mathrm{tor}}^{A,Q}\right],\\
\nabla_\phi L_{A,Q}-\nabla_\phi L_{\mathrm{all},I}
  &= \lambda_T\,\mathbb{E}_{z\sim\mu}
  \left[\nabla_\phi\left(\ell_{\mathrm{tor}}^{A,Q}-\ell_{\mathrm{tor}}^{\mathrm{all},I}\right)\right].
\end{aligned}
\label{eq:allocation_objective}
\end{equation}
Here, $A=\mathrm{all}$ and $Q=I$ denote broad intrinsic continuation. Because the corrupted-state exposure and non-torsional objectives are held fixed, changing $A$ or $Q$ changes the torsional contribution to the post-training gradient. The competing policies therefore differ in how a fixed corrective budget is allocated rather than in the states presented to the model.

The evaluation measures are separated because candidate generation and confidence-based pose selection are distinct stages in diffusion docking pipelines \citep{corso2023diffdock,corso2024deepconfident,zhu2024diffbindfr}. For $N$ complexes, $S$ generated poses, RMSDs $d_{ij}$, confidence scores $q_{ij}$, and RMSD threshold $a$, let
\[
d_{ij}=\operatorname{RMSD}(\widehat{x}_{ij},x_i^*).
\]
\begin{equation}
\begin{aligned}
S_{\mathrm{gen}}(a)
  &= \frac{1}{NS}\sum_{i,j}\mathbf{1}[d_{ij}<a],\\
S_{\mathrm{best}}(a)
  &= \frac{1}{N}\sum_i\mathbf{1}[\min_j d_{ij}<a],\\
j_i^*
  &= \arg\max_j q_{ij},
\qquad
S_{\mathrm{Top1}}(a)
  = \frac{1}{N}\sum_i\mathbf{1}[d_{i,j_i^*}<a].
\end{aligned}
\label{eq:endpoints}
\end{equation}
$(S_{\mathrm{gen}})$ measures generated-pose quality, $(S_{\mathrm{best}})$ measures sampling capacity, and $(S_{\mathrm{Top1}})$ measures confidence-selected utility \citep{corso2024deepconfident,zhu2024diffbindfr}, whereas fixed score fidelity is evaluated separately on common validation perturbations. Accordingly, better score fidelity does not imply better sampling capacity, and better sampling capacity does not imply a better confidence-selected pose. The primary $<5\,\text{\AA}$ threshold evaluates coarse pose recovery; supporting $<2\,\text{\AA}$ results assess stricter structural accuracy (Appendix~\ref{app:flexibility}). The relevant comparison in the next section is whether geometric information that identifies difficult pose states also points to the corrective supervision needed to improve them.

\section{Geometric Diagnosis and Corrective Action}
In these probes, low-rank tail energy is the squared Frobenius residual of the best rank-$k$ approximation to a smoothed ligand--receptor proximity matrix. Across the evaluated PDBBind and MOAD slices \citep{liu2015pdbbind,hu2005bindingmoad}, higher tail energy was associated with larger final RMSD and a greater prevalence of poses above $5\,\text{\AA}$ RMSD (Appendix Fig.~\ref{fig:tension_synthesis}a; Appendix Fig.~\ref{fig:lowrank_tail_energy}). The signal therefore identifies geometrically difficult corrupted poses. The motivation for testing both the statistic and its operational role is supported by equivariant-network work showing that geometric treatment can affect numerical stability and downstream physical error \citep{meng2024geonorm}. Its downstream meaning, however, depended on how it was operationalized. Direct ranking, noise-dependent gating, and contextual fusion did not preserve the same relationship with RMSD, and some gating transformations reversed the ordering induced by the raw statistic. Contextual fusion ranked highest in the aggregate descriptive comparison of the tested signal formulations. The corresponding method-family, cutoff, and weighting comparisons are reported in Appendix~\ref{app:geometric_diagnosis}.

The geometric rationale for focusing supervision is more direct at the torsional level. Internal ligand flexibility can be represented through torsional coordinates \citep{jing2022torsionaldiffusion,corso2023diffdock}, but equal angular errors do not imply equal Cartesian perturbations. For an atom $i$ at perpendicular distance $\rho_i$ from a torsional rotation axis, an angular error $\Delta\theta$ produces a chord displacement whose squared magnitude is
\begin{equation}
\begin{aligned}
\lVert\Delta \mathbf{x}_i\rVert^2
  &= 4\rho_i^2\sin^2\!\left(\frac{\Delta\theta}{2}\right)
  \approx \rho_i^2(\Delta\theta)^2,\\
\operatorname{RMSD}^2
  &\approx \Delta\boldsymbol{\theta}^{\mathsf T}
  G_{\mathrm{raw}}(\boldsymbol{\theta})\Delta\boldsymbol{\theta},
\qquad
G_{\mathrm{raw}}(\boldsymbol{\theta})
  = \frac{1}{n_{\mathrm{atom}}}J^{\mathsf T}PJ .
\end{aligned}
  \label{eq:cartesian_metric}
\end{equation}
The first relation gives the single-torsion consequence, while the local metric $G_{\mathrm{raw}}(\boldsymbol{\theta})$ extends the same principle across coupled torsional coordinates. Here, $J$ maps infinitesimal torsional changes into Cartesian displacements and $P$ removes rigid-body components. The derivation of the chord-displacement identity and the rigid-aligned pullback metric in Equation~\ref{eq:cartesian_metric}, together with its finite-difference check, is given in Appendix~\ref{app:torsion_priority}. Torsions can therefore differ in their Cartesian consequence even when their angular errors are comparable. This provides a geometric basis for testing focused torsional supervision, but it does not specify which corrective update the diffusion model should learn.

That limitation is visible in the directional probes. Reverse diffusion requires a directional denoising field over corrupted translational, rotational, and torsional states \citep{song2021scorebased,corso2023diffdock}. Geometry-derived finite-difference directions showed weak alignment with translation score targets and no consistent alignment with rotation score targets (Appendix Fig.~\ref{fig:tension_synthesis}c; Appendix Fig.~\ref{fig:rotation_signal_oracle}). Auxiliary geometry-loss gradients likewise showed weak or conflicting alignment with score-matching gradients across the evaluated parameter groups. An RMSD-improvement oracle and a score-aligned axis reference provide distinct analysis-only comparisons: the former uses immediate RMSD reduction, while the latter selects the tested signed axis with greatest agreement with the score target. Direct diagnostic directions were much less aligned with score targets than the score-informed axis reference. Full finite-difference, gradient-alignment, oracle-disagreement, and rotation-window analyses are reported in Appendix~\ref{app:diagnosis_action}. These probes motivate testing supervision allocation; they do not identify the cause of the trained policies' outcomes. The next section tests Cartesian consequence and conditional target consistency through torsional loss metrics and allocation masks; the probe directions are not used as training targets.

\section{Torsion Supervision Priorities and Study Design}
Conditional target consistency supplies a second basis for allocating torsional supervision. At noise level $s$, let $\mathbb E_{\mathrm{near}}$ and $\mathbb E_{\mathrm{random}}$ denote empirical averages over nearest-neighbor and random pairs of scalar torsion score targets for group $i$. Then
\begin{equation}
r_i(s)
  = \frac{\mathbb E_{\mathrm{near}}\!\left[|y-y'|\right]}
  {\mathbb E_{\mathrm{random}}\!\left[|y-y'|\right]},
\qquad
  c_i
  = \mathbf{1}\!\left[
  \max_{s\in\mathcal S} r_i(s)\leq 0.5
  \right].
\label{eq:consistency_proxy}
\end{equation}
The absolute differences are taken in score-target space without periodic angle wrapping; low $r_i(s)$ indicates that nearby represented states provide more similar targets than randomly paired states, and $c_i$ defines the edge-level training mask by requiring this relation across all probed noise levels. An edge is target-consistent when its neighboring targets are at most half as dissimilar as randomly paired targets at every probed noise level; under squared-error supervision, target variation conditional on a represented state contributes to conditional variance, so the measured ratio serves as a proxy for where corrections may be less consistently specified. Construction of $r_i(s)$, score-target pairing, noise-level aggregation, threshold selection, and mask generation are detailed in Appendix~\ref{app:torsion_priority}.
At the training-graph level, the consistency mask retained 61.2\% of PDBBind and 49.5\% of MOAD torsion edges; count-matched controls retained the same per-graph counts while changing assignments where possible.

The two allocation priorities were compared on 403 torsion-edge groups at four noise levels (1,612 time-indexed qualification rows), with Cartesian consequence measured by a static $90^\circ$ cross-pair displacement proxy and high consequence defined by a $1.0\,\text{\AA}$ threshold. This screening proxy is distinct from the pose-dependent matrix $G_{\mathrm{raw}}$; loss weighting uses its trace-normalized implementation $\widetilde G$. The resulting criteria overlapped but produced crossed assignments: 279 high-consequence rows were inconsistent, whereas 645 consistent rows had low Cartesian consequence (Appendix Fig.~\ref{fig:tension_synthesis}b). The full cross-tabulation is reported in Appendix~\ref{app:torsion_priority}. The underlying continuous measures had weak Pearson correlation (0.124). Geometry and target consistency therefore supply distinct allocation information rather than interchangeable rankings. Full threshold, fresh-bank, symmetry, periodicity, and qualification audits are reported in Appendix~\ref{app:torsion_priority}.

Each policy receives five additional training epochs from the same previously trained DiffDock model. A $3\times2$ design crosses allocation---all torsions, a count-matched subset, or the consistency mask---with the intrinsic angular metric $I$ or trace-normalized Cartesian metric $\widetilde G$ derived from Equation~\ref{eq:cartesian_metric}. All-$I$ is broad continuation, and the all-versus-consistent cells form the core $2\times2$ comparison. Count-matched controls preserve subset size while changing assignments where possible. Consistent versus Count-matched tests which torsions are selected; $\widetilde G$ versus $I$ tests how their errors are measured; differences between these effects test interaction. Policy definitions are given in Appendix~\ref{app:protocol}.

A final reconciliation policy combines broad intrinsic supervision with focused Cartesian supervision:
\begin{equation}
L_{\mathrm{mix}}
  = \frac{1}{2}L_{\mathrm{all},I}
  + \frac{1}{2}L_{\mathrm{consistent},\widetilde G}.
\label{eq:mixture_objective}
\end{equation}
The mixture is a separate seventh policy, not a factorial cell. Its exact half-weighted implementation and policy configuration are given in Appendix~\ref{app:protocol}, while the mask construction and qualification are detailed in Appendix~\ref{app:torsion_priority}. Having defined the competing priorities, the next section traces their consequences from the score objective through pose generation to confidence-based selection.

\section{Post-Training Results and Implications}
\begin{table}[t]
\centering
\scriptsize
\begin{tabular}{p{3.0cm}rrrr}
\toprule
Policy & Fixed loss & Generated $<5$ \AA & Best-of-10 $<5$ \AA & Top-1 $<5$ \AA \\
\midrule
Starting model & 1.6034 & 45.76 & 86.87 & 34.34 \\
Broad intrinsic & 1.6546 & 47.14 & 86.87 & 40.07 \\
Broad Cartesian & 1.6559 & 46.84 & 85.86 & 38.38 \\
Count-matched intrinsic & 1.6532 & 47.10 & 87.21 & 37.37 \\
Count-matched Cartesian & 1.6538 & 46.90 & 85.52 & 38.05 \\
Consistent intrinsic & 1.6519 & 46.84 & 85.86 & 38.38 \\
Consistent Cartesian & 1.6520 & 46.97 & 85.86 & 37.71 \\
Broad/focused mixture & 1.6529 & 46.94 & 86.20 & 39.06 \\
\bottomrule
\end{tabular}
\caption{Main post-training outcomes. Docking columns are percentages; fixed score loss is lower-is-better. The starting model is one checkpoint; trained-policy values are three-seed means.}
\label{tab:full_policy_summary}
\end{table}

Across three training seeds, Cartesian weighting, target-consistency allocation beyond count-matched controls, their factorial combination, and the separate broad/focused mixture showed no consistent advantage over broad continuation on the primary $<5\,\text{\AA}$ endpoints (Table~\ref{tab:full_policy_summary}; Appendix Fig.~\ref{fig:matched_results}). For the broad/focused mixture minus broad continuation, descriptive 95\% paired-complex bootstrap intervals were $[-0.61,\,0.17]$ percentage points for generated $<5\,\text{\AA}$ success and $[-4.38,\,2.36]$ percentage points for Top-1 success. The unaveraged starting checkpoint retained lower fixed score loss than the broad-continuation EMA state, which had higher mean generated-pose and Top-1 success. For broad minus starting, the corresponding descriptive paired-complex bootstrap intervals were $[0.17,\,2.56]$ and $[-1.68,\,13.47]$ percentage points (Table~\ref{tab:bootstrap_contrasts}); the Top-1 interval includes zero. Aggregate best-of-10 recovery was unchanged, but equal aggregate recovery does not establish that the same complexes contained successful candidates. This source-to-broad comparison includes checkpoint averaging and does not isolate an effect of continuation. At $<2\,\text{\AA}$, the mean Top-1 difference was $+0.67$ percentage points (Appendix Table~\ref{tab:two_angstrom_summary}).

Additional visual summaries and supporting details for the endpoint separation, seed-level policy contrasts, and flexibility changes are provided in Appendices~\ref{app:full_matched} and~\ref{app:flexibility}.

Ligand flexibility further separated these effects (Appendix Fig.~\ref{fig:matched_results}c). For complexes with $1$--$5$ torsions ($n=64$), the broad-continuation EMA state had higher mean generation, best-of-10 recovery, and Top-1 success than the starting checkpoint. For complexes with $6+$ torsions ($n=31$), mean generated success and best-of-10 recovery were lower while Top-1 success was higher. Broad continuation had the highest observed mean $<5\,\text{\AA}$ Top-1 success among trained policies, while the starting checkpoint retained the lowest fixed score loss and sampling differences varied with flexibility. Intervention activity and paired contrasts are reported in Appendices~\ref{app:intervention}--\ref{app:full_matched}; stratified values and evaluation-stability checks are reported in Appendices~\ref{app:flexibility}--\ref{app:robustness}.

\section{Discussion and Limitations}
Cartesian consequence and target consistency motivated focused supervision, but the tested policies did not consistently improve $<5\,\text{\AA}$ returned-pose success over broad continuation within five epochs. The directional probes show that identifying difficult states need not recover corrective directions; they do not establish the cause of the trained policies' outcomes. Generation and confidence-based selection remain distinct stages \citep{corso2023diffdock,corso2024deepconfident}. Top-1 measures returned-pose performance and compatibility with the fixed selector. For the starting and broad-continuation states, that selector performed below the expected uniform-random success rate at both reported thresholds (Appendix~\ref{app:flexibility}).

The study evaluates whole-pose outcomes without assigning failures to pocket placement or ligand conformation or measuring pairwise diversity among the ten generated poses; these separate analyses remain future work.

In higher-flexibility complexes, the broad EMA state had lower sampling recovery but higher mean Top-1 success than the starting checkpoint \citep{jing2022torsionaldiffusion,zhu2024diffbindfr}. RMSD and confidence selection do not establish physical validity or interaction recovery \citep{buttenschoen2023posebusters,errington2024interactionrecovery}; PoseBusters evaluation and external confirmation remain future work. Alternative allocation rules, adaptive weighting, longer post-training, or joint adaptation of generation and confidence selection could change the result; confidence-based feedback has coupled these stages in docking \citep{corso2024deepconfident}. Evidence from one architecture, 99 complexes, three training seeds, and one sampling seed leaves generalization to other models, datasets, and training budgets unresolved. The design problem is geometric supervision with broad score-field coverage \citep{song2021scorebased,corso2023diffdock}.

\begin{ack}
The authors gratefully acknowledge Niloofar Yousefi for her mentorship and support.
\end{ack}

\newpage
\bibliographystyle{plainnat}
\bibliography{references}

\clearpage
\appendix
\begin{figure*}[t]
\centering
\includegraphics[width=0.98\textwidth]{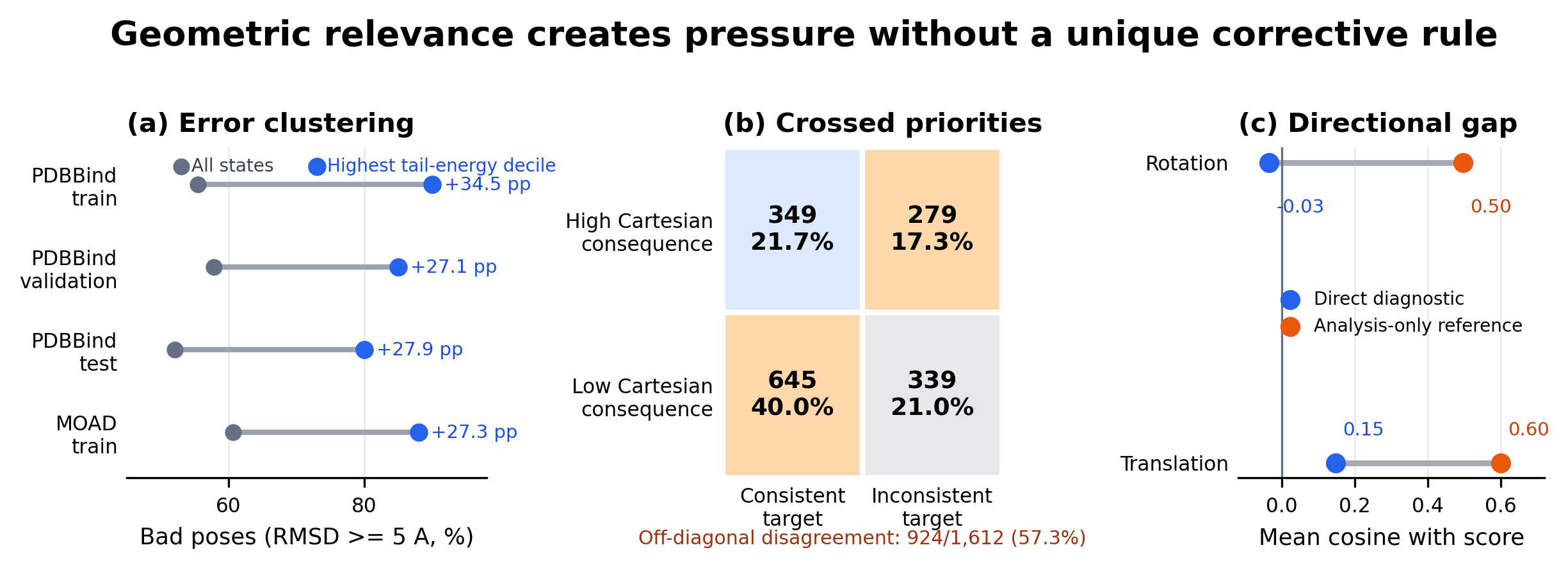}
\caption{(a) Bad-pose rates for all states and the highest tail-energy decile; (b) crossed Cartesian-consequence and target-consistency assignments; and (c) direct diagnostic and analysis-only reference alignment with the score target.}
\label{fig:tension_synthesis}
\end{figure*}

\section{Experimental Definitions and Fixed Evaluation}
\label{app:protocol}
This appendix defines the quantities, procedures, and fixed choices used in the analyses. The starting model is the inference-selected checkpoint from a 200-epoch full DiffDock score-model training run on the combined PDBBind, Binding MOAD, and PDBSidechain training stream. PDBSidechain is a project-derived preprocessing of protein structures from the RCSB Protein Data Bank source snapshot dated 2021-08-02 \citep{berman2000protein}. The starting model uses the six-layer equivariant graph architecture recorded for the run, with 32 scalar and 8 vector channels, second-order spherical-harmonic features, sinusoidal diffusion-time embeddings, and separate translation, rotation, and torsion score outputs. It is evaluated without additional score-model training and serves as the reference for every post-training policy.

\subsection{Policies and training stream}
The training stream contains PDBBind, MOAD, and PDBSidechain examples. The allocation rules act on PDBBind and MOAD torsion edges; PDBSidechain remains an all-torsion background in every trained policy. The same translation and rotation supervision is retained across policies. A torsion-edge group is the ordered list of rotatable ligand edges for one graph. If a graph has $m$ such edges, $e\in\mathbb{R}^{m}$ denotes the vector of sigma-normalized torsion prediction errors for that graph. Let $I_m$ be the $m\times m$ identity matrix. The trace-normalized local Cartesian matrix $\widetilde G$ is the training implementation of the pose-dependent matrix $G_{\mathrm{raw}}$ derived in Equation~\ref{eq:cartesian_metric}; its trace is rescaled to $m$ before it is used in a full-graph Cartesian loss. Let $C=\operatorname{diag}(c_1,\ldots,c_m)$ be the binary conditional-consistency mask and let $C_{\pi}=\Pi^{\mathsf T}C\Pi$ be the fixed within-graph count-matched permutation of that mask. The displayed quadratic forms below identify the error terms used by the policies. Their graph-level reductions divide by the number of supervised torsions, and masked Cartesian forms are trace-normalized after masking.

The starting model and seven trained-policy definitions are summarized in Table~\ref{tab:policy_definitions}.

\begin{table}[!h]
\centering
\scriptsize
\renewcommand{\arraystretch}{1.12}
\begin{tabular}{p{2.1cm}p{1.7cm}p{2.3cm}p{5.65cm}}
\toprule
Policy & Allocation & Torsional form or objective & Role in the comparison \\
\midrule
Starting model & none & none & Unchanged reference model \\
Broad intrinsic & all torsions & $I$ & Broad continuation \\
Broad Cartesian & all torsions & $\widetilde G$ & Cartesian weighting without selection \\
Count-matched intrinsic & permuted subset & $C_{\pi}$ & Generic sparse-allocation control \\
Count-matched Cartesian & permuted subset & $C_{\pi}\widetilde G C_{\pi}$ & Sparse Cartesian control \\
Consistent intrinsic & consistency mask $C$ & $C$ & Target-consistency allocation \\
Consistent Cartesian & consistency mask $C$ & $C\widetilde G C$ & Joint allocation and Cartesian weighting \\
Broad/focused mixture & broad plus $C$ & $\frac{1}{2}L_{\mathrm{all},I}+\frac{1}{2}L_{\mathrm{consistent},\widetilde G}$ & Reconciliation policy \\
\bottomrule
\end{tabular}
\caption{Definitions of the starting model and seven trained policies. The starting model is evaluated once; each trained policy is evaluated with three training seeds. The count-matched mask preserves the number of selected torsions within each graph while changing their assignments.}
\label{tab:policy_definitions}
\end{table}

For a policy with allocation $A$ and metric $Q$, the torsional contribution is represented by $\ell_{\mathrm{tor}}^{A,Q}$ in Equation~\ref{eq:allocation_objective}. The all-torsion intrinsic form is $e^{\mathsf T}I_me$; the all-torsion Cartesian form is $e^{\mathsf T}\widetilde G e$; and the two masked forms are $e^{\mathsf T}Ce$ and $e^{\mathsf T}C\widetilde GCe$, with $C_{\pi}$ substituted for $C$ in the count-matched policies. The starting model is the unchanged checkpoint against which the seven post-training policies are evaluated. The mixture is a separate policy: it combines the broad intrinsic objective with the consistent Cartesian objective after the factorial cells have been defined.

\subsection{Training and checkpoint rules}
Each trained policy uses five post-training epochs, corresponding to 13,000 optimizer steps, from the same starting model. The three training seeds are 20260812, 20260813, and 20260814. An exponential-moving-average (EMA) checkpoint is the running parameter average saved during training; the epoch-5 EMA checkpoint is the prespecified trained model for evaluation, with no later best-epoch selection. All 21 trained policy instances completed five epochs, saw 13,000 training batches, skipped zero training batches, and produced an epoch-5 EMA checkpoint. Within-policy objective changes document learning under each objective, but policy-specific training losses are not used as cross-policy evidence because the allocation and metric definitions change their scale and terms.

The recorded optimizer configuration uses Adam with learning rate $10^{-4}$, batch size 8, one gradient-accumulation step, zero weight decay, EMA rate 0.999, four data-loader workers, and deterministic training. No learning-rate scheduler or explicit learning-rate restart is configured.

The starting reference was evaluated using the original unaveraged inference-selected checkpoint. Each trained policy began from those source weights, used a newly initialized Adam optimizer, and was evaluated with the same prespecified epoch-5 EMA convention. Source-to-policy endpoint differences therefore include both parameter updates and checkpoint averaging, while trained-policy comparisons share the same optimizer initialization rule and EMA evaluation convention.

\subsection{Reproducibility and availability}
The software environment is specified by the accompanying \texttt{environment.yml} and \texttt{requirements.txt} declarations. The modified DiffDock implementation and analysis scripts used for training, evaluation, and appendix summaries are retained with the project materials, together with derived result summaries, split identifiers, and evaluation manifests. No verified hardware snapshot identifying the GPU model was captured in the experiment records. Raw PDBBind and MOAD data are not redistributed because access is governed by their licenses; the corresponding split identifiers, derived evidence, and provenance manifests are retained.

\subsection{Fixed evaluation}
Let $\mathcal A$ be the set of evaluation complexes, $N=|\mathcal A|=99$, and $S=10$ the number of generated poses per complex. For complex $i$, $\widehat{x}_{ij}$ is generated pose $j$, $x_i^*$ is the reference pose, $d_{ij}=\operatorname{RMSD}(\widehat{x}_{ij},x_i^*)$, and $q_{ij}$ is the confidence score assigned to that pose. The fixed-confidence Top-1 index is $j_i^*=\arg\max_jq_{ij}$. A complex is torsion-eligible when its ligand has at least one rotatable bond; the reported $0$, $1$--$5$, and $6+$ strata count those bonds. Generated-pose success, best-of-10 recovery, and Top-1 success use the definitions in Equation~\ref{eq:endpoints}.

The downstream evaluation uses the PDBBind time-split test set containing 99 complexes, ten generated poses per complex, sampling seed 20260820, and a schedule of 20 diffusion steps with 19 actual updates. A separately trained PDBBind confidence model, selected at its recorded epoch-75 checkpoint, ranks candidates for every score-model checkpoint and is not retrained during this comparison. The confidence model was trained separately using a different score-model distribution and was held fixed for the starting checkpoint and all post-training policies. Because the score models generate different candidate sets, a fixed confidence model can still select different poses; Top-1 therefore measures both returned-pose performance and compatibility between the candidate distribution and the selector. Fixed score fidelity is measured on the PDBBind no-ligand-overlap validation split using one shared perturbation bank generated with seed 20260820. A perturbation bank is a fixed collection of corrupted validation states, including their diffusion times, rigid perturbations, torsion perturbations, and score targets. The bank is reused for the starting model and every trained policy.

For checkpoint $\phi$, the common fixed score fidelity is
\begin{equation}
L_{\mathrm{fixed}}(\phi)
  = \frac{1}{|\mathcal F|}
    \sum_{\xi\in\mathcal F}\ell(\phi;\xi),
\label{eq:fixed_score_loss}
\end{equation}
where $\mathcal F$ is the shared set of corrupted validation states. The lower value is closer to the fixed reference score objective. It is not used as a substitute for the generated-pose, best-of-10, or Top-1 endpoints defined in Equation~\ref{eq:endpoints}.

The evaluation contract prespecified generated-pose and confidence-selected Top-1 success below $5\,\text{\AA}$ as co-primary endpoints, with $2\,\text{\AA}$ results retained as a stricter secondary analysis. Generated-pose success averages the fraction of the ten poses below this threshold; best-of-10 recovery records whether at least one pose per complex is below it; and Top-1 success records whether the pose selected by the fixed confidence model is below it. Conditional confidence-selection rate is the fraction of complexes for which Top-1 is below $5\,\text{\AA}$ among complexes with at least one generated pose below $5\,\text{\AA}$. The torsion-count strata and $5\,\text{\AA}$ conditional-selection analysis are supporting results in Appendix~\ref{app:flexibility}.

Policy means are averages over the three training seeds. The starting model has one evaluation because it has no training seed in this comparison. Paired-complex bootstrap intervals use 10,000 resamples and are descriptive; the three training seeds are used to assess direction consistency and effect size rather than to claim seed-level statistical significance. These intervals condition on the evaluated checkpoints and sampling seed; they resample complexes, not training or sampling seeds. They therefore do not quantify full uncertainty from retraining or resampling candidate sets. A two-percentage-point absolute docking-rate change with a matched seed direction was the prespecified materiality threshold.

\section{Geometric State Diagnosis}
\label{app:geometric_diagnosis}
For each corrupted state, the smoothed ligand--receptor proximity matrix is denoted by $H$. Its best rank-$k$ approximation is $H_k$, and the low-rank tail energy is
\[
T_k=\lVert H-H_k\rVert_F^2=\sum_{r>k}\sigma_r(H)^2.
\]
Here $\lVert\cdot\rVert_F$ is the Frobenius norm and $\sigma_r(H)$ is the $r$th singular value. The reported $k=1$ statistic is therefore the proximity-pattern energy left outside the leading component. A high-tail decile is the largest tenth of states ranked by $T_1$. The bad indicator in this appendix is final RMSD at least $5\,\text{\AA}$; the risk difference is the bad-pose rate in the high-tail decile minus the bad-pose rate in the full slice. Spearman correlation measures rank association between $T_k$ and final RMSD.

The slice-level tail-energy associations are summarized in Table~\ref{tab:low_rank_summary}.

\begin{table}[t]
\centering
\scriptsize
\begin{tabular}{lrrrrr}
\toprule
Slice & $n$ & Spearman $\rho$ & Baseline bad & High-tail bad & Risk difference \\
\midrule
PDBBind train & 1000 & 0.171 & 0.555 & 0.900 & 0.345 \\
PDBBind validation & 197 & 0.253 & 0.579 & 0.850 & 0.271 \\
PDBBind test, no receptor overlap & 48 & 0.164 & 0.521 & 0.800 & 0.279 \\
MOAD train & 1000 & 0.155 & 0.607 & 0.880 & 0.273 \\
\bottomrule
\end{tabular}
\caption{Definition and summary of the $k=1$ tail-energy association across the PDBBind train, validation, and test slices and MOAD train slice. These are diagnostic summaries, not trained-policy comparisons.}
\label{tab:low_rank_summary}
\end{table}

\begin{figure}[t]
\centering
\includegraphics[width=\linewidth]{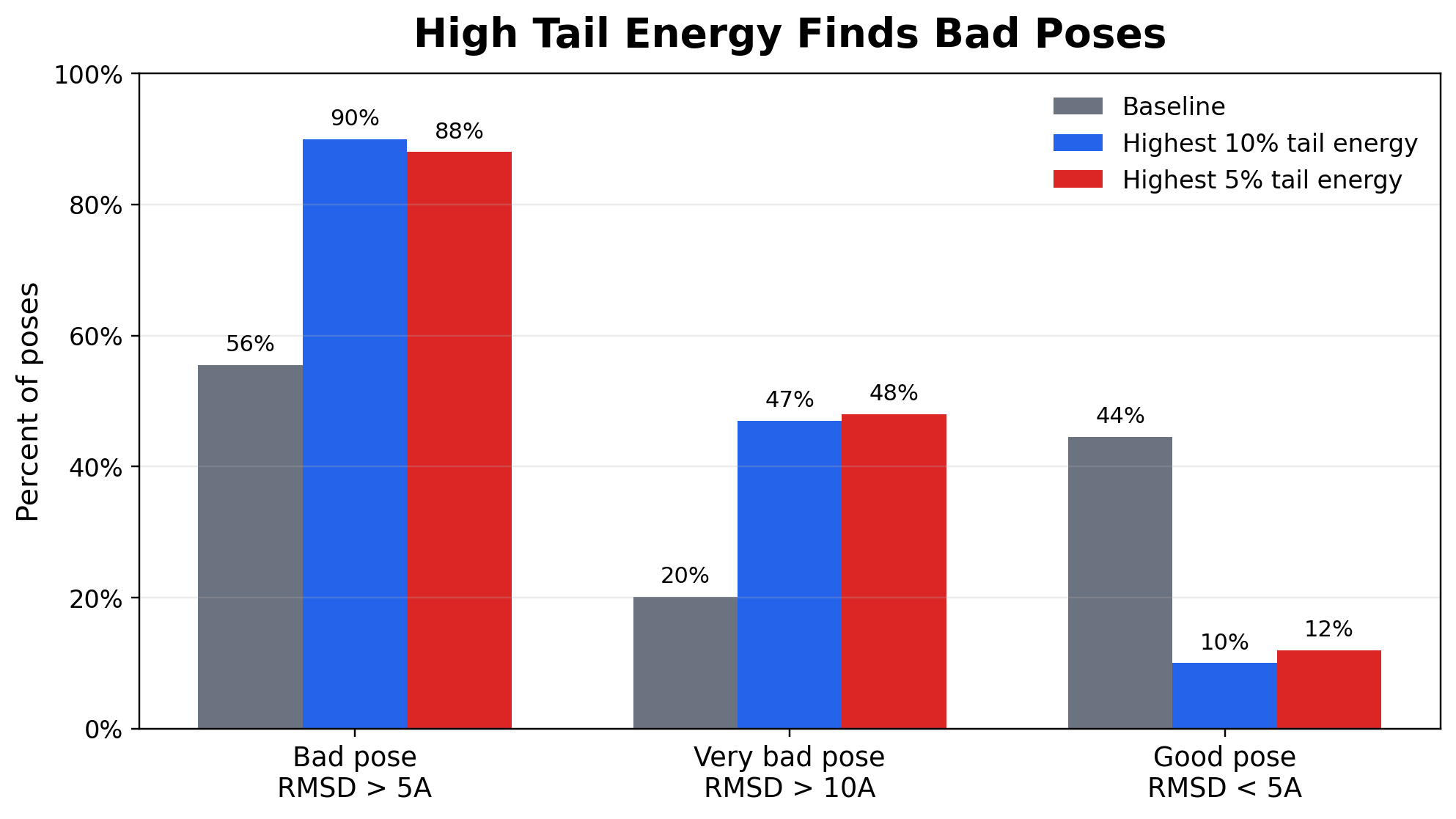}
\caption{Tail-energy association with difficult docking states in the 1,000-graph PDBBind training slice. Bars show pose-outcome rates across the full slice and within the highest 10\% and 5\% of tail energy. This diagnostic relationship does not establish a corrective training rule.}
\label{fig:lowrank_tail_energy}
\end{figure}

The method-family analysis compares three uses of the same geometric signal: standalone tail-energy ranking, noise-dependent gating, and contextual fusion. A method view is one candidate statistic or transformation evaluated on the same probe rows. Candidate percentiles summarize the relative position of a view within the tested candidate set; tie-aware ranking preserves ties rather than treating them as independent evidence. The 11 contextual-fusion, five sigma-restricted, and two standalone candidates were evaluated as correlated views. Their method-family and weighting summaries are descriptive supporting analyses; representative signal-role values are shown in Table~\ref{tab:signal_role_numeric}.

\begin{table}[t]
\centering
\scriptsize
\begin{tabular}{lrrrr}
\toprule
Signal use & PDBBind train & PDBBind val & PDBBind test & MOAD train \\
\midrule
Standalone tail energy & 0.171 & 0.253 & 0.164 & 0.155 \\
Contextual fusion & 0.238 & 0.293 & 0.127 & 0.233 \\
Translation-gated fusion & $-0.252$ & $-0.255$ & $-0.318$ & $-0.213$ \\
Rotation-gated fusion & $-0.230$ & $-0.230$ & $-0.308$ & $-0.187$ \\
\bottomrule
\end{tabular}
\caption{Representative Spearman associations between signal variants and final RMSD. The two noise-dependent gated variants reverse the positive association observed for standalone tail energy and contextual fusion across the four evaluated slices.}
\label{tab:signal_role_numeric}
\end{table}

\section{Directional and Gradient Diagnostics}
\label{app:diagnosis_action}
The directional probes evaluated one corrupted state after small perturbations along a finite candidate-direction set. For the rotation probes, the 26-direction set contained the signed coordinate axes, face diagonals, and cube diagonals. For candidate direction $b$, the probe recorded the change in the geometric signal, alignment with the score target and model prediction, and the resulting RMSD. If $d_0$ is the RMSD after the unmodified probe and $d_b$ is the RMSD after candidate $b$, the pure RMSD reference selected
\begin{equation}
 b_{\mathrm{RMSD}} = \arg\max_b\Delta d_b,
 \qquad
 \Delta d_b=d_0-d_b.
\label{eq:oracle_direction}
\end{equation}
For vectors $u$ and $v$, directional alignment is $\cos(u,v)=u^{\mathsf T}v/(\lVert u\rVert\lVert v\rVert)$ when both norms are nonzero. The pure RMSD reference uses the crystal pose only to identify the tested direction with the greatest immediate geometric improvement; it is not a trainable target and is unavailable to the docking model during prediction. A rank-selected direction is the candidate with the largest observed decrease in the diagnostic statistic. The score-aligned axis reference selects among tested signed axes using agreement with the score target; it is the analysis-only reference used for the alignment rows in Table~\ref{tab:diagnosis_action_summary} and Panel (c) of Figure~\ref{fig:tension_synthesis}. A support-constrained oracle retains candidates passing the rank-energy and target-alignment screens, then selects the candidate maximizing the product of diagnostic decrease and nonnegative cosine with the score target. The support-constrained choice was compared with the rank-selected diagnostic direction, while $b_{\mathrm{RMSD}}$ served as the pure RMSD reference. A free-sign reference is a post-hoc upper-bound comparison that permits the sign of a tested axis to be chosen using the score-target direction.

The oracle probe skipped the final singleton batch of the 197-graph PDBBind validation slice, leaving 196 graphs. In this probe using the 26-direction set, the rank-selected and support-constrained choices agreed in 2 cases (1.0\%), and a support-constrained choice was available in 97 cases (49.5\%). Within those available cases, the support-constrained choice produced a positive immediate RMSD change in 56.7\% of cases, compared with 51.5\% for the rank-selected choice. These values describe local directional probes rather than full docking performance; their role is to show that the geometric diagnostic did not reliably recover a single pose-improving direction. The measured alignment summary is given in Table~\ref{tab:diagnosis_action_summary}; candidate-level direction records and the additional gradient and rotation-window summaries are retained as supporting audit material.

\begin{table}[t]
\centering
\small
\begin{tabular}{llrrr}
\toprule
Direction choice & Component & $n$ & Mean cosine & Positive fraction \\
\midrule
Direct finite difference & Translation & 1,574 & 0.147 & 59.6\% \\
Direct finite difference & Rotation & 1,548 & $-0.035$ & 46.1\% \\
Energy-ranked & Translation & 1,574 & 0.118 & 59.8\% \\
Energy-ranked & Rotation & 1,548 & $-0.048$ & 46.3\% \\
Score-aligned axis reference & Translation & 1,574 & 0.599 & 92.2\% \\
Score-aligned axis reference & Rotation & 1,548 & 0.496 & 84.7\% \\
Free-sign reference & Translation & 1,574 & 0.835 & 100.0\% \\
Free-sign reference & Rotation & 1,548 & 0.831 & 100.0\% \\
\bottomrule
\end{tabular}
\caption{Alignment of tested direction choices with the corresponding score target. The diagnostic choices remain weakly aligned, whereas the score-aligned axis and free-sign rows use analysis-only information and serve as references rather than deployable corrective rules. The pure RMSD reference is defined separately.}
\label{tab:diagnosis_action_summary}
\end{table}

\begin{figure}[t]
\centering
\includegraphics[width=\linewidth]{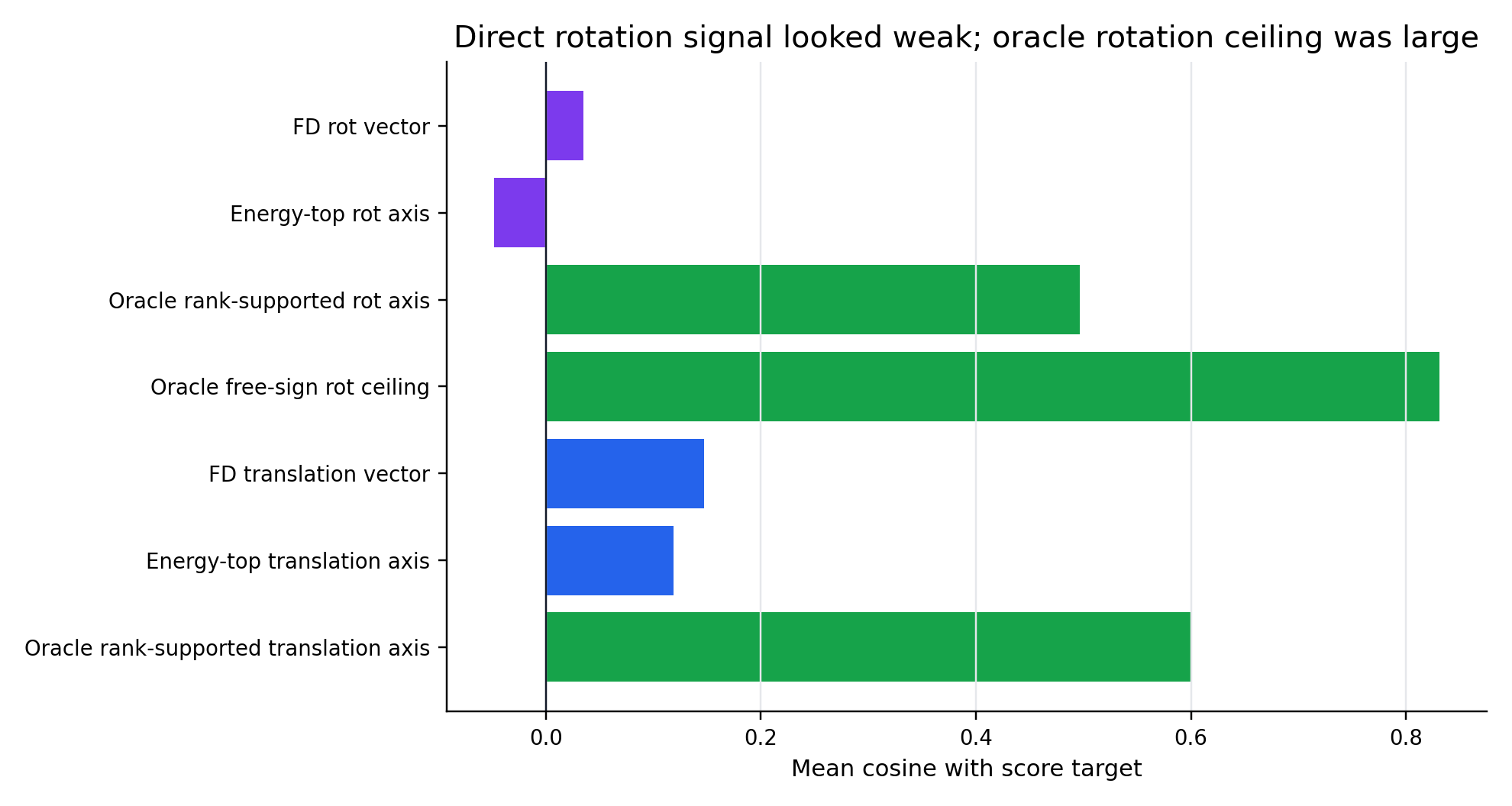}
\caption{Directional diagnostic and analysis-only reference comparison. Direct diagnostic directions are compared with the score-aligned axis reference used in the alignment summary. The analysis-only reference is unavailable to the docking model and is used to assess diagnostic-to-action agreement.}
\label{fig:rotation_signal_oracle}
\end{figure}

The gradient table reports cosine alignment between the gradient of a gated geometric objective and the gradient of a score objective. A parameter group is the global model, the translation head, the rotation head, the torsion head, or the shared and remaining parameters. Rows with no valid batches are not interpreted as zero alignment. The rotation-window analysis partitions the single fusion-oracle trajectory into early and late epoch windows; its quantities are descriptive trajectory summaries rather than replicated training results. Representative gradient values are shown in Table~\ref{tab:gradient_alignment_summary}.

\begin{table}[t]
\centering
\scriptsize
\begin{tabular}{l l r r r}
\toprule
Gate cutoff & Reference score loss & Global cosine & Rotation-head cosine & Batches \\
\midrule
2.0 & Rotation component & 0.026 & $-0.064$ & 20 \\
2.0 & Full score loss & $-0.044$ & $-0.064$ & 20 \\
3.0 & Rotation component & 0.026 & $-0.158$ & 20 \\
3.0 & Full score loss & $-0.009$ & $-0.158$ & 20 \\
\bottomrule
\end{tabular}
\caption{Representative mean cosine alignments between a gated geometric objective and score losses over 20 valid batches. Near-zero global values and negative rotation-head values show weak or conflicting alignment; rows with no valid batches are omitted.}
\label{tab:gradient_alignment_summary}
\end{table}

\section{Torsion Priority Qualification}
\label{app:torsion_priority}
The qualification set contained 403 torsion-edge groups evaluated at four noise levels $s\in\{0.1,0.3,0.6,0.9\}$. For edge $i$, let $\mathcal U_i$ be atoms moved by the torsional rotation and $\mathcal V_i$ the atoms left fixed.

\subsection{Derivation of the Cartesian metric}
For a point at perpendicular distance $\rho_i$ from a torsional axis, write its position relative to a point on the axis as $u_i+v_i$, where $u_i$ is parallel to the axis and $v_i$ is perpendicular. A rotation by $\Delta\theta$ leaves $u_i$ unchanged and gives
\[
\lVert x_i'-x_i\rVert^2
 = \lVert R(\Delta\theta)v_i-v_i\rVert^2
 = 2\lVert v_i\rVert^2\bigl(1-\cos\Delta\theta\bigr)
 = 4\rho_i^2\sin^2\!\left(\frac{\Delta\theta}{2}\right).
\]
For small $\Delta\theta$, this is $\rho_i^2(\Delta\theta)^2+O((\Delta\theta)^4)$, which gives the first relation in Equation~\ref{eq:cartesian_metric}. For simultaneous small changes $\Delta\boldsymbol{\theta}$, the Cartesian displacement satisfies
\[
\Delta\mathbf{x}
 = J\,\Delta\boldsymbol{\theta}
 + O\!\left(\lVert\Delta\boldsymbol{\theta}\rVert^2\right),
\qquad
\operatorname{RMSD}^2
 = \frac{1}{n_{\mathrm{atom}}}
 \lVert P\Delta\mathbf{x}\rVert^2,
\]
where $J$ is the Cartesian Jacobian with one torsional column per edge and $P$ is the symmetric idempotent projector that removes local rigid translation and rotation components. Substitution and $P^{\mathsf T}P=P$ yield
\[
\operatorname{RMSD}^2
 \approx \frac{1}{n_{\mathrm{atom}}}
 \Delta\boldsymbol{\theta}^{\mathsf T}J^{\mathsf T}PJ
 \Delta\boldsymbol{\theta}
 = \Delta\boldsymbol{\theta}^{\mathsf T}
 G_{\mathrm{raw}}(\boldsymbol{\theta})\Delta\boldsymbol{\theta}.
\]
The diagonal entries of $G_{\mathrm{raw}}$ encode individual torsional consequence, while off-diagonal entries encode coupled Cartesian displacement. The implementation trace-normalizes this matrix to form $\widetilde G$ before applying the graph-level training loss, as defined in Appendix~\ref{app:protocol}; the qualification calculation uses the unnormalized geometric form above. Finite-difference validation over 13,248 direction tests gave a 90th-percentile relative error of $3.2\times10^{-4}$ for the analytic local form and $3.6\times10^{-3}$ for the full local approximation. The off-diagonal Frobenius contribution reached $0.70$ at the 90th percentile, so the coupled terms are retained rather than replaced by diagonal weights.

The static cross-pair displacement proxy rotates $\mathcal U_i$ by $\pi/2$ around the edge axis and computes
\[
d_{\mathrm{cross}}(i)=
\left[
\frac{1}{|\mathcal U_i||\mathcal V_i|}
\sum_{u\in\mathcal U_i}\sum_{v\in\mathcal V_i}
\left(\lVert x'_u-x'_v\rVert-\lVert x_u-x_v\rVert\right)^2
\right]^{1/2}.
\]
High Cartesian consequence means $d_{\mathrm{cross}}(i)\geq1.0\,\text{\AA}$. This static screen is used only for the qualification cross-tabulation; it is distinct from the pose-dependent matrix $G_{\mathrm{raw}}$; loss weighting uses $\widetilde G$.

For the consistency calculation, the numerator in Equation~\ref{eq:consistency_proxy} is the mean absolute difference between each score target and its nearest-neighbor target; the denominator is the mean absolute difference over randomly paired score targets. Periodic angle wrapping is used for angular perturbation handling elsewhere, but not for this score-target ratio. A row is consistent when $r_i(s)\leq0.5$; the edge-level training mask $c_i$ is one only when this holds at every probed noise level. Table~\ref{tab:priority_cross} counts time-indexed $(i,s)$ rows, not final edge-level mask values.

\begin{table}[t]
\centering
\small
\begin{tabular}{lrr}
\toprule
 & Consistent row & Inconsistent row \\
\midrule
High Cartesian consequence & 349 & 279 \\
Low Cartesian consequence & 645 & 339 \\
\bottomrule
\end{tabular}
\caption{Crossed supervision priorities among 1,612 time-indexed qualification observations from 403 torsion-edge groups and four noise levels. Rows use the static $90^\circ$ cross-pair screen and the score-target consistency threshold defined in the text.}
\label{tab:priority_cross}
\end{table}

The off-diagonal cells show the crossed assignments between Cartesian consequence and target consistency. The continuous measures have Pearson correlation $0.124$. Fresh-bank prediction uses $k$ nearest represented states or a local-$5$ estimator to predict torsional targets; lower prediction error indicates better agreement with the observed target. Symmetry strata use the fraction of duplicate atoms on the rotated side, with low symmetry at most $0.25$ and elevated symmetry above $0.25$. Periodic-wrapping diagnostics compare raw and wrapped angular updates. The allocation audit reports the selected edge fraction as selected edges divided by all torsion edges, while lookup failure means that a runtime graph could not be matched to its fixed allocation key.

The metric coverage and finite-difference checks are summarized in Table~\ref{tab:metric_qualification}.

\begin{table}[t]
\centering
\scriptsize
\begin{tabular}{lrrrrr}
\toprule
Source & Processed & Multi-torsion & Metric states & Direction tests & Graph failures \\
\midrule
PDBBind & 192 & 173 & 1656 & 13248 & 0 \\
\bottomrule
\end{tabular}
\caption{Coverage and finite-difference checks for the local Cartesian metric on the PDBBind qualification set. These checks qualify the metric approximation and do not establish downstream training benefit.}
\label{tab:metric_qualification}
\end{table}

At the qualification-set level, the 90-degree screen yielded the cross-tabulation above. At the full training-graph level, the locked $0.5$ consistency threshold selected 61.2\% of PDBBind torsion edges and 49.5\% of MOAD torsion edges, with zero allocation lookup failures. On the independent perturbation bank, label agreement was 85.3\% for PDBBind and 90.8\% for MOAD; Cohen's kappa was 0.692 and 0.814, respectively. These quantities qualify the allocation rule and do not establish that it improves docking.

\section{Intervention Activity}
\label{app:intervention}
An intervention is active when it changes the torsional loss term or allocation while the run completes the specified training exposure. Within-policy objective response is the change from the first to the last logged training epoch for the same objective; it is not a cross-policy ranking. For a parameter group $g$, let $\Delta\phi_g^{(a)}$ and $\Delta\phi_g^{(b)}$ be flattened checkpoint changes from the common starting model. Update similarity is
\[
\operatorname{cos}_{g}(a,b)=
\frac{\left(\Delta\phi_g^{(a)}\right)^{\mathsf T}\Delta\phi_g^{(b)}}
\lVert\Delta\phi_g^{(a)}\rVert\lVert\Delta\phi_g^{(b)}\rVert.
\]
The training-integrity records show five epochs, 13,000 batches, zero skipped batches, and an epoch-5 EMA checkpoint for every trained instance. Every policy's own training objective declined over its logged five-epoch trajectory. The metric ablation provides an independent activity check: aligned-versus-permuted update cosine was approximately $0.35$ in the torsion head, compared with approximately $0.99$ in the translation and rotation heads. Thus the metric change redirected the torsional part of the update even when the non-torsional heads changed similarly.

\section{Full Matched Post-Training Study}
\label{app:full_matched}
The matched-study results contain one starting checkpoint and seven trained policies. A policy mean is the arithmetic mean across its three training seeds; the accompanying standard deviation is the sample standard deviation across those seeds. The starting row has no training-seed standard deviation because it is one fixed checkpoint. A paired seed contrast is policy A minus policy B for the same training seed. A paired-complex contrast uses the same generated complex and subtracts the two endpoint indicators before bootstrapping across complexes.

The primary practical contrast is the broad/focused mixture minus broad intrinsic continuation. Its generated $<5\,\text{\AA}$ differences by seed were $-0.30$, $0.00$, and $-0.30$ percentage points; its Top-1 differences were $-4.04$, $+2.02$, and $-1.01$ points. The primary reference contrast is broad intrinsic continuation minus the starting model: generated success changed by $+0.91$, $+2.32$, and $+0.91$ points, while Top-1 changed by $+10.10$, $+5.05$, and $+2.02$ points. The same paired arithmetic was applied to every predeclared threshold, endpoint, stratum, and policy pair, with 10,000-resample descriptive bootstrap intervals.

Relative to the starting checkpoint, the broad-intrinsic mean fixed loss increased by $0.0512$, comprising $+0.0292$ in translation loss, $+0.0224$ in rotation loss, and $-0.0004$ in torsion loss. The records therefore do not identify the increase as a torsion-specific effect or attribute it to the learning rate, optimizer state, schedule, or checkpoint averaging.

\begin{table}[t]
\centering
\scriptsize
\begin{tabular}{@{}lccc@{}}
\toprule
Contrast & Generated $<5\,\text{\AA}$ & Best-of-10 $<5\,\text{\AA}$ & Top-1 $<5\,\text{\AA}$ \\
\midrule
Broad $-$ starting & $+1.38\ [0.17,\,2.56]$ & $+0.00\ [-4.38,\,4.38]$ & $+5.72\ [-1.68,\,13.47]$ \\
All-G $-$ All-I & $-0.30\ [-0.71,\,0.10]$ & $-1.01\ [-2.69,\,0.34]$ & $-1.68\ [-5.39,\,1.68]$ \\
Count-G $-$ Count-I & $-0.20\ [-0.61,\,0.17]$ & $-1.68\ [-3.70,\,-0.34]$ & $+0.67\ [-3.03,\,4.04]$ \\
Consistent-I $-$ Count-I & $-0.27\ [-0.61,\,0.03]$ & $-1.35\ [-2.69,\,-0.34]$ & $+1.01\ [-3.03,\,5.05]$ \\
Consistent-G $-$ Count-G & $+0.07\ [-0.40,\,0.54]$ & $+0.34\ [-0.67,\,1.68]$ & $-0.34\ [-3.03,\,2.36]$ \\
Consistent-G $-$ Consistent-I & $+0.13\ [-0.24,\,0.51]$ & $+0.00\ [-1.01,\,1.01]$ & $-0.67\ [-4.04,\,2.69]$ \\
Mixture $-$ All-I & $-0.20\ [-0.61,\,0.17]$ & $-0.67\ [-1.68,\,0.00]$ & $-1.01\ [-4.38,\,2.36]$ \\
\bottomrule
\end{tabular}
\caption{Descriptive 95\% paired-complex percentile bootstrap intervals for the predeclared all-complex contrasts. Entries are percentage-point differences, with the first policy minus the second; intervals use 10,000 resamples and are not seed-level significance intervals.}
\label{tab:bootstrap_contrasts}
\end{table}

\begin{figure*}[t]
\centering
\includegraphics[width=0.98\textwidth]{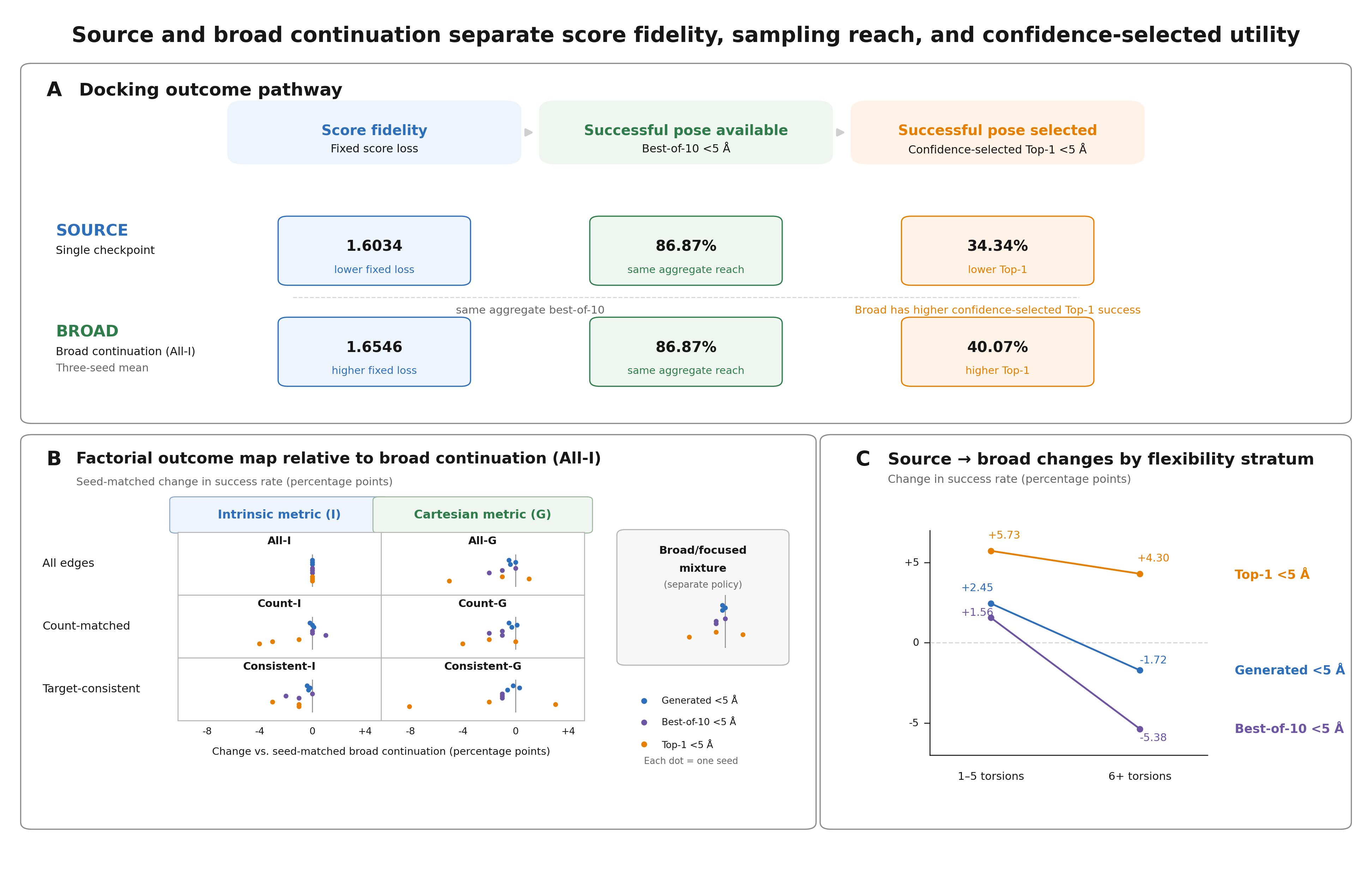}
\caption{Integrated summary of the post-training endpoint separation. (a) The starting model and broad intrinsic continuation differ in fixed score loss, aggregate best-of-10 recovery, and confidence-selected Top-1 success; the starting model is one checkpoint and the broad values are three-seed means. (b) Seed-matched changes for the factorial policies are shown relative to broad intrinsic continuation; each dot is one training seed, and the broad/focused mixture is shown separately because it is not a factorial cell. (c) Source-to-broad changes are shown by torsional flexibility stratum. The figure re-expresses the quantities reported in the matched-study and flexibility summaries.}
\label{fig:posttraining_outcome_separation}
\end{figure*}

\begin{figure*}[t]
\centering
\includegraphics[width=0.98\textwidth]{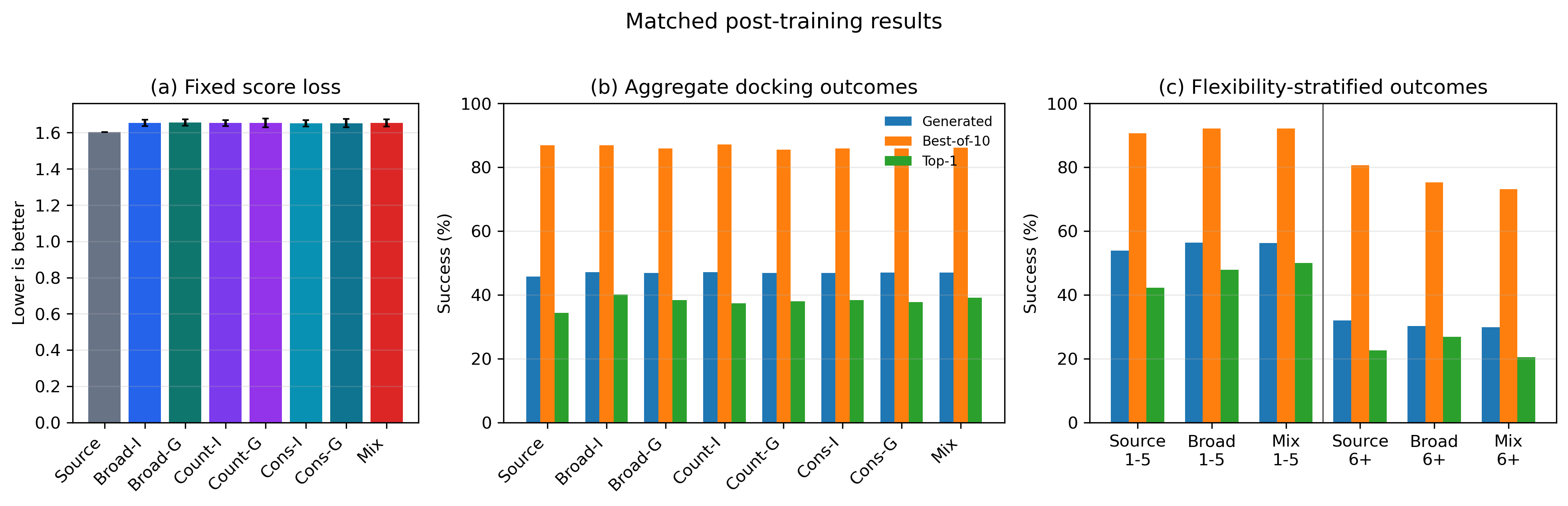}
\caption{Matched post-training outcomes. (a) Fixed score loss; (b) aggregate generated-pose, best-of-10, and confidence-selected Top-1 success; and (c) the same docking endpoints for the $1$--$5$ and $6+$ torsion strata. Bars show the starting model and three-seed policy means; error bars show seed standard deviations where available. The common PDBBind evaluation uses ten poses per complex and a $5\,\text{\AA}$ threshold.}
\label{fig:matched_results}
\end{figure*}

The integrated endpoint and seed-matched outcome summaries are shown in Figures~\ref{fig:posttraining_outcome_separation} and~\ref{fig:matched_results}.

\section{Sampling, Confidence Selection, and Flexibility}
\label{app:flexibility}
The flexibility strata are defined by the number of rotatable ligand bonds: zero, $1$--$5$, and $6+$; torsion-eligible is the union of the latter two strata. Generated success is averaged over poses, best-of-10 is averaged over complexes, and Top-1 is averaged over confidence-selected complexes, as defined in Equation~\ref{eq:endpoints}. All threshold variants use the same generated poses and confidence scores; only the RMSD indicator threshold changes.

The $5\,\text{\AA}$ flexibility-stratified endpoints are summarized in Table~\ref{tab:flexibility_summary}.

\begin{table}[t]
\centering
\scriptsize
\begin{tabular}{llrrrr}
\toprule
Policy & Stratum & $n$ & Generated & Best-of-10 & Top-1 \\
\midrule
Starting model & all & 99 & 45.76 & 86.87 & 34.34 \\
Broad intrinsic & all & 99 & 47.14 & 86.87 & 40.07 \\
Mixture & all & 99 & 46.94 & 86.20 & 39.06 \\
Starting model & $1$--$5$ torsions & 64 & 53.91 & 90.63 & 42.19 \\
Broad intrinsic & $1$--$5$ torsions & 64 & 56.35 & 92.19 & 47.92 \\
Mixture & $1$--$5$ torsions & 64 & 56.20 & 92.19 & 50.00 \\
Starting model & $6+$ torsions & 31 & 31.94 & 80.65 & 22.58 \\
Broad intrinsic & $6+$ torsions & 31 & 30.22 & 75.27 & 26.88 \\
Mixture & $6+$ torsions & 31 & 29.89 & 73.12 & 20.43 \\
\bottomrule
\end{tabular}
\caption{Flexibility-stratified percentages on the 99-complex PDBBind test evaluation at the $5\,\text{\AA}$ threshold. Values use ten generated poses per complex; the four zero-torsion complexes are retained in the all-complex result but are not interpreted as a subgroup.}
\label{tab:flexibility_summary}
\end{table}

Supporting $2\,\text{\AA}$ endpoints for all policies are listed in Table~\ref{tab:two_angstrom_summary}.

\begin{table}[t]
\centering
\scriptsize
\begin{tabular}{@{}p{2.2cm}rrr@{}}
\toprule
Policy & Generated $<2\,\text{\AA}$ & Best-of-10 $<2\,\text{\AA}$ & Top-1 $<2\,\text{\AA}$ \\
\midrule
Starting model & 17.98 & 49.49 & 14.14 \\
Broad intrinsic & 18.99 & 48.82 & 14.81 \\
Broad Cartesian & 19.12 & 49.16 & 15.15 \\
Count-matched intrinsic & 18.89 & 49.16 & 14.14 \\
Count-matched Cartesian & 19.02 & 49.49 & 15.49 \\
Consistent intrinsic & 18.92 & 49.49 & 15.15 \\
Consistent Cartesian & 18.96 & 49.49 & 14.81 \\
Broad/focused mixture & 19.16 & 49.49 & 14.81 \\
\bottomrule
\end{tabular}
\caption{Supporting $2\,\text{\AA}$ docking endpoints on the 99-complex PDBBind test evaluation. Values are percentages; trained-policy values are three-seed means. The $5\,\text{\AA}$ threshold remains primary.}
\label{tab:two_angstrom_summary}
\end{table}

For each complex with at least one generated pose below $5\,\text{\AA}$, conditional confidence-selection rate records whether the confidence-selected pose is also below $5\,\text{\AA}$. It measures selection among recoverable complexes, not the probability of generating a recoverable pose. In the high-flexibility stratum, broad continuation reduced best-of-10 recovery by 5.38 points relative to the starting model while increasing Top-1 by 4.30 points; the mixture was 6.45 points below broad continuation in Top-1.

For a uniform random choice among the $S$ generated poses, the expected success for complex $i$ is
\begin{equation}
\mathbb{E}_{J\sim\operatorname{Uniform}\{1,\ldots,S\}}
\mathbf{1}[d_{iJ}<a]
  = \frac{1}{S}\sum_{j=1}^{S}\mathbf{1}[d_{ij}<a].
\label{eq:random_reference}
\end{equation}
Averaging over complexes gives exactly $S_{\mathrm{gen}}(a)$, so every generated-pose value is also the corresponding uniform-random reference. The resulting source-versus-broad confidence-selection gaps at both reported thresholds are shown in Table~\ref{tab:random_selection_reference}.

\begin{table}[t]
\centering
\scriptsize
\begin{tabular}{@{}llrrr@{}}
\toprule
Policy & Threshold & Random reference & Confidence Top-1 & Top-1 minus reference \\
\midrule
Starting model & $2\,\text{\AA}$ & 17.98 & 14.14 & $-3.84$ \\
Broad intrinsic & $2\,\text{\AA}$ & 18.99 & 14.81 & $-4.18$ \\
Starting model & $5\,\text{\AA}$ & 45.76 & 34.34 & $-11.42$ \\
Broad intrinsic & $5\,\text{\AA}$ & 47.14 & 40.07 & $-7.07$ \\
\bottomrule
\end{tabular}
\caption{Confidence-selected success compared with the expected success from uniformly selecting one of ten generated poses. Values are percentages; the random reference equals generated-pose success.}
\label{tab:random_selection_reference}
\end{table}

Original DiffDock reported 35.0\% Top-1 success below $2\,\text{\AA}$ with ten generated poses and 61.7\% below $5\,\text{\AA}$ on its 363-complex PDBBind test set \citep{corso2023diffdock}. The present starting model reached 14.14\% and 34.34\%, respectively, while broad continuation reached 14.81\% and 40.07\%. These values are not a direct benchmark because the test-set size, score-model training run, model capacity, and evaluation setup differ.

\section{Robustness and Evaluation Stability}
\label{app:robustness}
An evaluation outlier is a generated pose with RMSD greater than $1{,}000\,\text{\AA}$. Across the 22 evaluated checkpoints, 63 of 21,780 poses exceeded this value and occurred in eight complexes. The complete-case sensitivity analysis removes every complex associated with an outlier from every policy before recomputing the paired contrasts. After this removal, the mixture-minus-broad differences were $-0.26$ points for generated $<5\,\text{\AA}$ and $-1.10$ points for Top-1. The numerical failures therefore do not account for the primary policy conclusion.

Training-time monitoring validation resamples diffusion time and pose perturbations when the validation examples are transformed. A validation bank replay holds the checkpoint fixed and compares two independently sampled collections of corrupted states. In the replay, the mean score loss changed from 1.690 to 1.508, the median from 1.318 to 1.166, and the mean diffusion time changed from 0.475 to 0.519. This is a bank-difficulty comparison, not evidence of parameter improvement. The fixed validation bank used for the final policy comparison is separate and common to all checkpoints.

The isolated convolution replay examines one complex under a normal and a failing corrupted pose. A layer value is the maximum absolute activation recorded at that module. The failing pose increased the pose-readout maximum from 2.65 to $5.18\times10^{17}$, with progressive amplification through the convolution stack; the source checkpoint also failed on that perturbation. This is an inherited numerical limitation and is not assigned as the cause of the aggregate RMSD result.

\end{document}